\documentclass[fleqn,10pt]{wlscirep}
\usepackage[utf8]{inputenc}
\usepackage[T1]{fontenc}

\usepackage{graphicx}%
\usepackage{multirow}%
\usepackage{amsmath,amssymb,amsfonts}%
\usepackage{mathtools}
\usepackage{amsthm}%
\usepackage{mathrsfs}%
\usepackage{xcolor}%
\usepackage{textcomp}%
\usepackage{algorithm}%
\usepackage{algorithmicx}%
\usepackage{algpseudocode}%
\usepackage{float}
\usepackage{subcaption}
\usepackage{color,soul}

\title{An Explainable Transformer Model for Alzheimer’s Disease Detection Using Retinal Imaging}

\author[1]{Saeed Jamshidiha} 

\affil[1]{
Nanotechnology, Biotechnology, Information Technology and Cognitive Science Laboratory, University of Tehran,Tehran, Iran}

\author[2, *]{Alireza Rezaee} 

\affil[2]{Department of Mechatronics, School of Intelligent Systems, College of Interdisciplinary Science and Technology, University of Tehran, Tehran,Iran}

\author[3]{Farshid Hajati} 

\affil[3]{School of Science and Technology, Faculty of Science, Agriculture, Business and Law, University of New England, Parramatta, NSW 2150, Australia}

\author[4]{Mojtaba Golzan}
\affil[4]{Graduate School of Health, University of Technology Sydney, Ultimo, NSW 2007, Australia}

\author[3,5]{Raymond Chiong}
\affil[5]{School of Medicine and Public Health, College of Health, Medicine and Wellbeing, The University of Newcastle, Callaghan, NSW 2308, Australia \newline}

\affil[*]{Corresponding Author: arrezaee@ut.ac.ir}

\begin{abstract}
Alzheimer's disease (AD) is a neurodegenerative disorder that affects millions worldwide. In the absence of effective treatment options, early diagnosis is crucial for initiating management strategies to delay disease onset and slow down its progression. In this study, we propose Retformer, a novel transformer-based architecture for detecting AD using retinal imaging modalities, leveraging the power of transformers and explainable artificial intelligence. The Retformer model is trained on datasets of different modalities of retinal images from patients with AD and age-matched healthy controls, enabling it to learn complex patterns and relationships between image features and disease diagnosis.
To provide insights into the decision-making process of our model, we employ the Gradient-weighted Class Activation Mapping algorithm to visualize the feature importance maps, highlighting the regions of the retinal images that contribute most significantly to the classification outcome. These findings are compared to existing clinical studies on detecting AD using retinal biomarkers, allowing us to identify the most important features for AD detection in each imaging modality.
The Retformer model outperforms a variety of benchmark algorithms across different performance metrics by margins of up to 11\%.
\end{abstract}

\begin{document}


\flushbottom
\maketitle
%
%
\thispagestyle{empty}

\section*{Introduction}
Alzheimer's disease (AD) is a prevalent neurodegenerative disorder characterized by cognitive decline, memory loss, and impaired daily functioning \cite{alzheimer20162016}.
Early diagnosis of AD can enable the management of modifiable risk factors (such as lifestyle changes), significantly delaying the onset and progression of the disease \cite{sperling2011testing, dubois2016preclinical, sperling2011toward}.
A reduction in macular volume and the thickness of the retinal nerve fiber layer (RNFL) is observed in AD patients by Iseri et al. \cite{iseri2006relationship}. RNFL thickness is assessed and observed by Kirbas et al. to be significantly less in patients with AD and without visual impairment than in healthy age-matched controls \cite{kirbas2013retinal}. A decrease in RNFL thickness in mild to moderate stages of AD is also confirmed by Kromer et al. \cite{kromer2014detection}. A decrease in retinal thickness in mildly cognitively impaired patients and in AD patients is observed by Den et al., confirming the link between neurodegenerative diseases and retinal changes \cite{den2017retinal}. Furthermore, Cipollini et al. observe that OCT images can show changes in RNFL thickness, as well as a reduction in macular volume in the early stages of AD \cite{cipollini2020neurocognitive}.  Shin et al. report a reduction in retinal microvasculature, but not in the retinal thickness of AD patients, and suggest OCT-angiography (OCTA) of retinal microvasculature for detection of this reduction \cite{shin2021changes}.
A relationship is also suggested between changes in the thickness of the choroid layer and AD by Cheung et al. \cite{cheung2021retinal, cheung2017imaging}.
The meta-analysis by Ge et al. \cite{ge2021retinal} of the research published on the retinal biomarkers of AD reveals the thickness of peripapillary RNFL, total macular volume, and the thickness of the subfoveal choroid layer were significantly reduced in comparison with healthy controls (HCs) ($p<0.001$ in all cases), pointing to these as the most important retinal biomarkers of AD.
To summarise, the RNFL thickness, macular volume, the thickness of the choroid layer, and the overall retinal thickness are significant retinal biomarkers of AD.

The use of artificial intelligence (AI) in the diagnosis of AD, as well as various forms of retinopathy, including AD-related retinpoathy is an active area of research \cite{zhou2023foundation, badar2020application, ashayeri2024retinal, ebrahimighahnavieh2020deep, ebrahimi2021deep, ebrahimi2019transfer}. A modular system, using machine learning (ML) components is proposed by Tian et al. to diagnose AD using retinal images, achieving an accuracy of 82.44\% \cite{tian2021modular}. This modular design consists of an ensemble of neural networks to filter out low-quality images, a separate neural network for vessel segmentation, and a support vector machine (SVM) for the final classification. 
EfficientNet-B2 \cite{tan2019efficientnet} is used by Cheung et al. to train, validate, and test a model to diagnose AD with an accuracy of 83.6\% \cite{cheung2022deep}. 
Using a convolutional neural network (CNN) based on the ResNet architecture \cite{he2016deep}, the method presented by Wisely et al. achieves a recall of 72.7\% and a precision of 74.4\% on a dataset of multimodal images \cite{wisely2022convolutional}. 
Wang et al. compare the performances of different ML models, including K-nearest neighbours (KNN), random forest, and logistic regression, among others, to diagnose AD using features obtained from OCT images, reporting accuracies between 67\% and 74\% \cite{wang2022machine} . 
The method proposed by Corbin et al. utilises EfficientNet-B3 \cite{tan2019efficientnet} to explain 22.4\% of the variance in cognitive scores using coloured fundus images and metadata, while explaining 9.3\% of the variance using fundus data alone \cite{corbin2022assessment}.
Using VGG-16 \cite{simonyan2014very}, the method presented by Yousefzadeh et al. can classify AD with an accuracy of 71.4\%, and by leveraging explainable AI (XAI), it can assess the continuum of AD. This method validates the retinal vasculature as a biomarker for AD \cite{yousefzadeh2024neuron}.
The method presented by Kim et al. utilises the MobileNet \cite{sinha2019thin} architecture to classify AD using retinal fundus images, achieving an area under the curve (AUC) of the receiver operating characteristic (ROC) of 0.927 \cite{kim2024efficient}.

A number of pretrained foundation models (PFMs) have been proposed to detect various forms of retinopathy. RETFound is a foundation model pretrained on unlabeled retinal images, and an OCT version and a fundus version of it have been published \cite{zhou2023foundation}. FLAIR is a multimodal vision-language model trained on images of the retina and descriptions of the features and pathologies of the retina in the form of text prompts, thus encoding expert knowledge along with the image data \cite{silva2023foundation}. RET-CLIP is another multimodal vision-language model trained on fundus images and medical diagnoses \cite{du2024ret}. 

While different computer vision methods have been used to diagnose AD using retinal biomarkers, there is room for improvement in performance. While recent studies \cite{cheung2022deep, cheung2021retinal, vij2022systematic} have demonstrated the potential of deep learning models in detecting AD from retinal images, these approaches often rely on complex architectures and may not provide clear insights into the decision-making process.
However, medical image datasets are usually smaller than those commonly used in general machine vision tasks.
As a result, special care should be taken when designing neural network architectures for medical use cases.

There has been an exponential growth in the medical applications of vision transformers \cite{shamshad2023transformers, hussain2025effresnet}. The applications of vision transformers in medical image classification include COVID-19 classification, tumor classification, classification of retinal diseases \cite{shamshad2023transformers}, fracture prediction in X-ray images \cite{alam2024integrated}.
To apply transformers to small-scale multi-modal datasets, Dai et al. \cite{dai2021transmed} combine CNNs and transformers, and apply the methodology proposed by Touvron et al. \cite{touvron2021training} for the data-efficient image transformer (DEiT), which involves knowledge distillation.
To alleviate the challenges of working with small-sized expert-annotated datasets in the context of X-ray disease diagnosis using vision and language models, Monajatipoor et al. \cite{monajatipoor2022berthop} combine an unsupervised vision encoder with a language encoder pretrained on text-only data to classify 14 common thoracic chest diseases.
In order to train a transformer to perform image segmentation in the context of medical image segmentation, the method proposed by Valanarasu et al. \cite{valanarasu2021medical} proposes a gated axial attention mechanism that can work well with small datasets.
To summarise, the commonly-employed methods to adapt transformers to medical machine vision problems with small datasets include using convolutional filters, using modified attention mechanisms, knowledge distillation, and using pretrained models.

With the increasing application of AI in different use cases, including medical settings \cite{hussain2025dcssga, asim2022circ, khan2025robust}, the need for explainable AI (XAI) is becoming increasingly apparent \cite{dwivedi2023explainable, hekmat2025differential}. Since more advanced ML methods, such as deep learning, are inherently opaque, a number of different approaches have been proposed to make these methods' decision-making more transparent \cite{dwivedi2023explainable, selvaraju2017grad}. 
While more advanced computer vision methods such as transformers \cite{vaswani2017attention, dosovitskiy2020image} are expected to perform better than classical ML methods, since medical applications of AI especially require safety and transparency as they concern the lives and the health of human beings, deep learning methods are less frequently utilised in practice in this context.

In this paper, we propose Retformer, a novel transformer-based architecture to detect AD in retinal images, including visible light fundus images and OCT scans. This approach offers the promise of significantly increasing detection accuracy compared to the more common CNN-based architectures such as EfficientNet, VGG, and ResNet used for machine vision. The Retformer model is designed to be able to work with relatively small datasets. As a result, the Retformer model uses rotary position embedding (RoPE) \cite{su2024roformer}, convolutional layers, grouped query attention \cite{ainslie2023gqa}, and swish-gated linear units (SwiGLU) \cite{shazeer2020glu}.
In order to alleviate the problem with the lack of transparency of deep neural networks, we leverage an XAI framework, namely the Gradient-weighted Class Activation Mapping (Grad-CAM) algorithm \cite{selvaraju2017grad}, to explore why the method we propose makes the decisions it makes, and how it aligns with clinical studies and medical knowledge. The advantage of this approach over the more common method of visualising the attention heads of the transformer is that the Grad-CAM algorithm is able to visualise any layer in the network, while attention visualisation is only possible on the attention heads. Using the Grad-CAM algorithm, we are able to probe the feature map at the penultimate layer of the network, right before the classifier head.

Pre-training is not employed in this research, and the Retformer model is trained from scratch using the retinal images, as these images are different from images of everyday objects found in the datasets commonly used in machine vision.

The major contributions of this paper can be summarised as follows:
\begin{itemize}
	\item Proposing a novel transformer-based model for retinal images
	\item Utilising the proposed model for AD detection using retinal images.
	\item Using the XAI framework to obtain clinical-grade classifications.
	\item Exploring the most important features for AD detection in different retinal imaging modalities.
	\item Leveraging XAI to compare the model's classifications with clinical studies.
\end{itemize}

\section{Materials and Methods}
\label{sec.mat}
In this research, different retinal image modalities (i.e. fundus images and cross-sectional OCT scans) are utilised to detect AD. For this purpose, retinal images have been obtained from open datasets \cite{dryad, fundus}.
To train the models, select the optimal set of hyperparameters, and compare the performances of the models, we have utilised nested cross-validation, where the data is repeatedly split into different train, validation and test subsets, and the results are averaged over all folds. This process is explained in detail in the following subsections.

\subsection{Data}
The retinal image modalities used in this study include visible light fundus images and OCT images. Visible light fundus images are obtained using imaging devices that include an ophthalmic lens, a camera, and a flashlight. These images are used to diagnose various retinopathies, including glaucoma, age-related macular degeneration, diabetic retinopathies, and optic nerve disorders. OCT images are obtained using light reflections, and show multiple cross-section of the retina, thereby containing volume-related information.

OCT data were obtained from the open dataset published by Bissig et al.\cite{dryad}. This dataset includes 224 three-dimensional OCT images (1,120 two-dimensional slices) from 14 AD patients and 14 age-matched HCs, along with OCT images from 8 young adults, which we did not use in this research. The data we used was perfectly balanced, with 50\% AD images and 50\% HCs. The exclusion of young adults from the OCT dataset is aimed at minimising the impact of age-related features on the neural network's decision-making. By focusing exclusively on older adults, we can ensure that the model learns to recognise age-independent biomarkers indicative of Alzheimer's disease rather than age-related artifacts. This approach allows us to isolate the relevant features associated with the disease.

Fundus images were obtained from TRENDv2 \cite{fundus} and 1000 fundus images \cite{cen2021automatic}. The TRENDv2 dataset includes fundus images from 20 patients with chronic diseases (controlled hypertension, controlled type-2 diabetes mellitus, AD) and 8 healthy age-matched controls \cite{fundus}. The 1000 fundus images dataset includes 1000 fundus images with 39 different retinopathies \cite{cen2021automatic}. Since the size of TRENDv2 is not enough to train deep models, we utilised the HCs from this dataset to augment the data. The resulting aggregated dataset is imbalanced, with 14\% AD images versus 86\% HCs. 

\begin{figure}[H]
	\centering
	\scalebox{1}[2]{
		\includegraphics[angle=90, scale=.6]{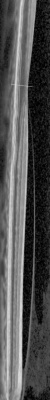}
	}
	\caption{A slice of an OCT image from an AD patient}
	\label{fig.OCT_slice}
\end{figure}

To enable data visualisation and model explanation, and augment the data, the three-dimensional OCT images were converted into two-dimensional slices. These slices are shown in Figure \ref{fig.OCT_slice}. 

The preprocessing steps undertaken in this study have been kept to a minimum to allow the models to learn the optimal features for image classification in an end-to-end manner. 
The preprocessing steps of the fundus dataset include resizing the images to 50x50 pixels, and rescaling the pixel values from the [0, 255] range to the [0, 1] range. In order to balance the class distribution of the images, we employed random oversampling of the minority class (AD) on the training subset of the images in our nested cross-validation scheme.
The preprocessing steps of the OCT dataset include extracting two-dimensional slices from the three-dimensional OCT images, resizing the two-dimensional slices to 50x50 pixels, and rescaling the pixel values to the [0, 1] range. Since this dataset is balanced, no oversampling was employed for the OCT images.

\subsection{Retformer}
\begin{figure}[h]
	\centering
	\includegraphics[scale=.1]{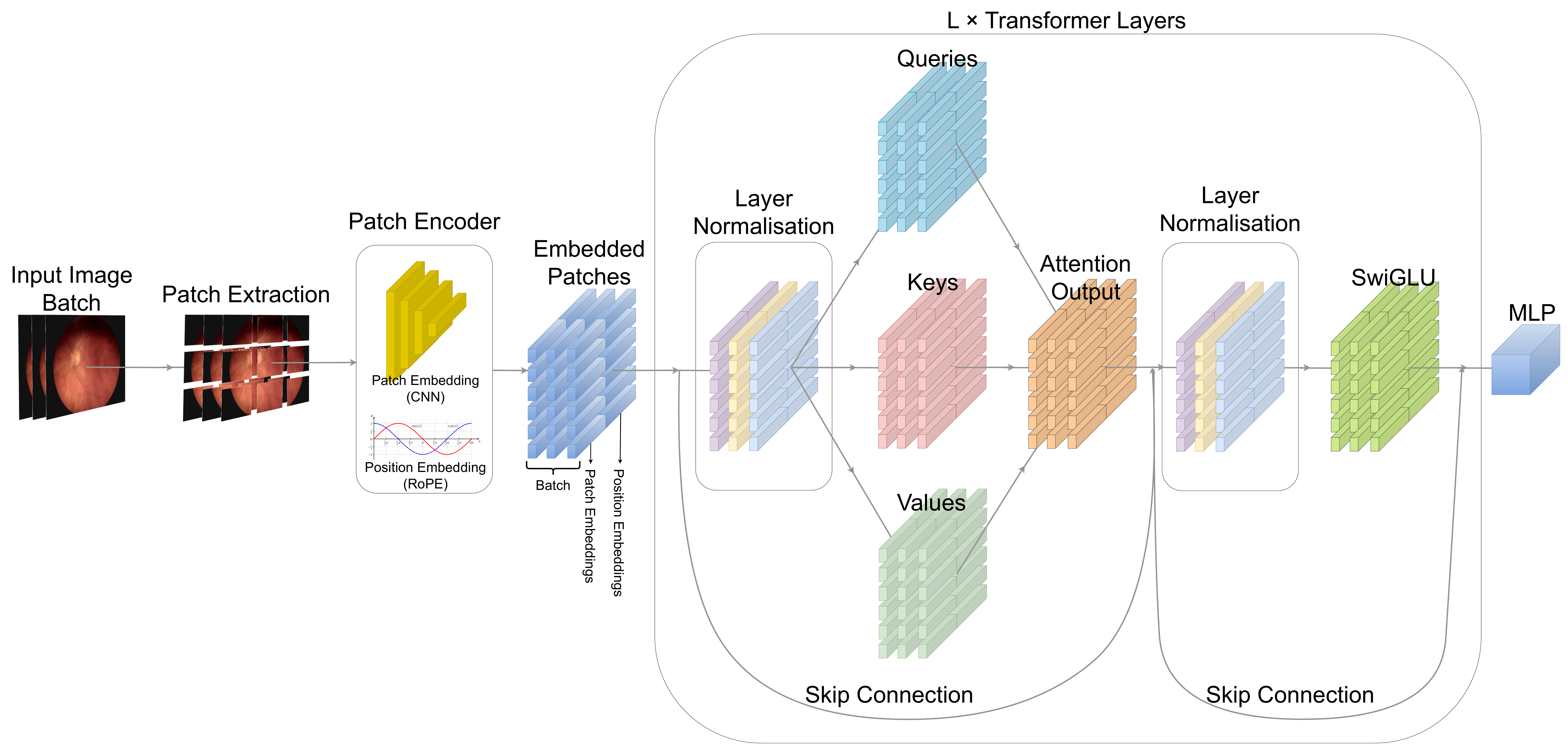}
	\caption{The Retformer model}
	\label{fig.vit}
\end{figure}

We propose Retformer, a transformer-based model to classify different modalities of retinal images into AD patients and HCs. The Retformer model is an encoder-only transformer that converts input images into a compact latent representation. This representation allows images to be classified using a classifier head, typically a multi-layer perceptron (MLP), which makes predictions based on the encoded features. The proposed architecture can be seen in Figure \ref{fig.vit}.

Given the relatively small size of typical medical image datasets in comparison with general machine vision datasets, each component of the Retformer model has been carefully chosen to mitigate the risk of overfitting, ensuring reliable performance even when dealing with small datasets. As a result, we use a non-trainable position encoding scheme, and convolutional layers to embed the input image to the transformer representation space, and use grouped query attention (GQA) to lower the number of model parameters while maintaining performance \cite{ainslie2023gqa}.

The Retformer model consists of three key architectural components: an input embedding layer, multiple transformer layers, and a multilayer perceptron (MLP) classifier head.

\begin{itemize}
    \item \textbf{Input Embedding Layer:} This layer includes patch extraction (dividing images into patches), a CNN-based learnable patch encoder, and a non-learnable rotary position encoder (RoPE) \cite{su2024roformer}. Using a CNN-based patch embedding improves robustness to overfitting compared to MLP-based embeddings, which is critical when training on small-scale medical datasets. The non-learnable RoPE further reduces the risk of overfitting by avoiding additional trainable parameters.
    
    \item \textbf{Transformer Layers:} These layers incorporate layer normalization, grouped query attention (GQA), SwiGLU activation functions, and two skip connections. GQA replaces standard multihead attention to limit overfitting while preserving performance, and SwiGLU activation has been shown to outperform GELU in small datasets, enhancing representation learning efficiency.
    
    \item \textbf{MLP Classifier Head:} The final component translates the learned representations into predictions.
\end{itemize}

Overall, this design specifically targets the challenges of small-scale medical datasets by reducing overfitting risks and improving generalization without compromising model capacity.

In the Retformer model, the input image is first transformed into a sequence of one-dimensional patches through a convolutional layer, which maps the two-dimensional image into a $D_E$-dimensional space. To take into account the spatial relationships between these patches, $D_E$-dimensional RoPE position embeddings are added to them. The position embeddings used in this model are one-dimensional, as experiments have demonstrated that two-dimensional embeddings do not provide any performance improvement for vision transformers \cite{dosovitskiy2020image}.The input embedding results in a matrix of size $N_P \times D_E$, where $N_P$ is the number of patches.

Each transformer layer consists of layer normalisation, a GQA head \cite{ainslie2023gqa}, a residual connection between the layer input and the output of the GQA head, another layer normalisation, a SwiGLU activation \cite{shazeer2020glu}, and a residual connection from the output of the GQA head to the layer output. The scaled dot-product attention head \cite{vaswani2017attention}, which is the building block of GQA, is defined using query matrix $\mathbf{Q}$, key matrix $\mathbf{K}$, and value matrix $\mathbf{V}$, which are trainable $D_A$ dimensional linear projections of the input matrix. Denoting the input matrix to transformer layer $l$ with $\mathbf{P_l}$, and the trainable projections used with attention head $i$ with $\mathbf{W_l^{q,i}}$, $\mathbf{W_l^{k,i}}$, and $\mathbf{W_l^{v,i}}$, these vectors can be defined as follows:

\begin{equation}
	\begin{aligned}
		\mathbf{Q^i_l} = \mathbf{W_l^{q,i}} \mathbf{P_l}, \\
		\mathbf{K^i_l} = \mathbf{W_l^{k,i}} \mathbf{P_l}, \\
		\mathbf{V^i_l} = \mathbf{W_l^{v,i}} \mathbf{P_l}.
	\end{aligned}
\end{equation}

The scaled dot-product attention head $i$ in layer $l$ is then defined as \cite{vaswani2017attention}:

\begin{equation}
	A(\mathbf{Q^i_l}, \mathbf{K^i_l}, \mathbf{V^i_l}) = s\Big(\frac{\mathbf{Q^i_l}\mathbf{K^i_l}^T}{\sqrt{D_A}}\Big)\mathbf{V^i_l},
\end{equation}
where $T$ denotes vector transposition and $s(\cdot)$ denotes the softmax function.

The GQA utilises multiple attention heads in parallel, where queries are grouped, and keys and values are shared among each group. The outputs of these heads are concatenated:

\begin{equation}
	\mathbf{G_l} = \Big[	
	\underbrace{
		A(\mathbf{Q^1_l}, \mathbf{K^1_l}, \mathbf{V^1_l}) \mathbin\Vert A(\mathbf{Q^2_l}, \mathbf{K^1_l}, \mathbf{V^1_l}) \mathbin\Vert 
	}_{\text{Attention group 1}}
	\cdots 
	\underbrace{
		\mathbin\Vert 
		A(\mathbf{Q^H_l}, \mathbf{K^g_l}, \mathbf{V^g_l})
	}_{\text{Attention group g}}
	\Big]_{N_P \times D_AH},
\end{equation}
where $\mathbin\Vert$ denotes concatenation, $H$ is the number of attention heads, and $g$ is the number of query groups. Denoting the output of the GQA in layer $l$ with $\mathbf{G_l}=[m_j], j \in [1, D_AH]$, layer normalisation $n(\mathbf{G_l})$ can be obtained from \cite{ba2016layernormalization}:

\begin{equation}
	n(m_j) = \frac{m_j-\mu}{\sigma},
\end{equation}
where $n(m_j)$ is the normalised output, and $\mu$ and $\sigma$ are the mean and standard deviation of the elements of $\mathbf{G_l}$:

\begin{equation}
	\begin{aligned}
		\mu = \frac{1}{D_AH}\sum_{j=1}^{D_AH}m_j, \\
		\sigma = \sqrt{\frac{1}{D_AH}\sum_{j=1}^{D_AH}\Big(m_j-\mu\Big)}.
	\end{aligned}
\end{equation}

The SwiGLU function is defined as \cite{shazeer2020glu}:

\begin{equation}
	f(\mathbf{x}) = swish(\mathbf{x}\mathbf{W_s}+\mathbf{B_s})\otimes(\mathbf{x}\mathbf{V_s}+\mathbf{C_s}),
\end{equation}
where $\mathbf{W_s}$ and $\mathbf{V_s}$ are SwiGLU's learnable weights and $\mathbf{B_s}$ and $\mathbf{C_s}$ are its learnable biases, $\otimes$ is element-wise multiplication, and:

\begin{equation}
	swish(x) = \frac{x}{1+e^{-x}}.
\end{equation}

The output of the transformer layer can be formulated as follows:
\begin{equation}
	y_l = f(n(\mathbf{M_l})) + \mathbf{M_l} + \mathbf{P_l},
\end{equation}
where $y_l$ is the output of the $l$-th transformer layer.

RoPE is a non-learnable position embedding method that encodes relative position instead of absolute position, i.e. the dot-product of the query and key vectors depends on the relative position of the query vector relative to the key vector, and not on the absolute positions of each of the query and key vectors \cite{su2024roformer}. Denoting the patch embedding of the i-th input patch with $x_i$,
	\begin{equation}
		\mathbf{Q^i_1}\mathbf{K^j_1}^T = g(x_i, x_j, i-j),
	\end{equation}
	where $g(.)$ is the dot-product of the query and key vectors, and it is a function of the patch embeddings of the i-th and j-th patches, and their relative position $i-j$.
	
	The RoPE embedding achieves this by using rotation to encode position information. In the case of a 2-dimensional embedding space \cite{su2024roformer}:
	\begin{equation}
		\begin{aligned}
			\mathbf{Q^i_1} = \mathbf{W_1^{q,1}} \mathbf{x_i} e^{\sqrt{-1}i\theta}, \\
			\mathbf{K^i_1} = \mathbf{W_1^{k,1}} \mathbf{x_j} e^{\sqrt{-1}j\theta}, 
		\end{aligned}
	\end{equation}
	where $\theta$ is a fixed angle. As a result,
	
	\begin{equation}
		\mathbf{Q^i_1}\mathbf{K^j_1}^T = \Re\{(\mathbf{W_1^{q,1}} \mathbf{x_i})(\mathbf{W_1^{k,1}} \mathbf{x_j})^* e^{\sqrt{-1}(i-j)\theta}\},
	\end{equation}
	where $\Re$ represents the real part, and $^*$ represents complex conjugation.
	
	For $D_E > 2$, RoPE is defined as \cite{su2024roformer}:
	
	\begin{equation}
		\begin{aligned}
			\mathbf{Q^i_1} = \mathbf{R_{\theta, i}^{D_E}} \mathbf{W_1^{q,1}} \mathbf{x_i}, \\
			\mathbf{K^i_1} = \mathbf{R_{\theta, j}^{D_E}} \mathbf{W_1^{k,1}} \mathbf{x_j},
		\end{aligned}
	\end{equation}
	
	\begin{equation}
		\mathbf{R_{\theta, i}^{D_E}} = 
		\begin{bmatrix*}
			\cos i\theta_1, & -\sin i\theta_1, & 0, & 0, & \cdots, & 0, & 0 \\
			\sin i\theta_1, & \cos i\theta_1, & 0, & 0, & \cdots, & 0, & 0 \\
			0, & 0, & \cos i\theta_2, & -\sin i\theta_2, & \cdots, & 0, & 0 \\
			0, & 0, & \sin i\theta_2, & \cos i\theta_2, & \cdots, & 0, & 0 \\
			\vdots, & \vdots, & \vdots, & \vdots, & \cdots, & \vdots, & \vdots \\
			0, & 0, & 0, & 0, & \cdots, & \cos i\theta_{\frac{D_E}{2}}, & -\sin i\theta_{\frac{D_E}{2}} \\
			0, & 0, & 0, & 0, & \cdots, & \sin i\theta_{\frac{D_E}{2}}, & \cos i\theta_{\frac{D_E}{2}}
		\end{bmatrix*},
	\end{equation}
	where $\theta_i$ is set to \cite{su2024roformer}:
	\begin{equation}
		\theta_i = 10000^{\frac{-2(i-1)}{D_E}}, \quad i \in [1, 2, \cdots, \frac{D_E}{2}].
	\end{equation}
	
	Using RoPE instead of a learnable position embedding decreases the number of learnable parameters of the Retformer model, making it robust to overfitting. Furthermore, the relative position encoding scheme of RoPE is equivalent to location invariance in the context of image patch embedding, since it only encodes the relative position of the patches and not their absolute locations.

This study does not use pre-trained models; instead, it trains the Retformer model from scratch using retinal images, as these retinal images differ from the everyday objects typically found in commonly used machine vision datasets.

\subsection{Hyperparameter Tuning}
\label{ssec.hyperparameters}
To obtain an unbiased estimation of the performance of any ML model, a three-way split is required between train, validation, and test datasets. However, when the data is randomly split into train, validation, and test sets, it is possible that the test set contains samples that the model can classify more accurately. 

In order to decrease the random effects of the two-way split between train and test sets, K-fold cross-validation splits the data into k folds, and on each iteration, assigns one fold to the test set, and the remaining k-1 folds to the train set. Nested cross-validation is an extension of K-fold cross-validation that employs an inner loop and an outer loop to produce a three-way split between train, validation and test sets. The data is split into $K_1$ folds, and one fold is assigned to the test set and $K_1 - 1$ folds are assigned to the train and validation sets. These folds are then split into $K_2$ folds, and one fold is assigned to the validation set while the remaining folds are assigned to the train set. As a result of this procedure, a more accurate assessment of the model's performance can be obtained.

\subsubsection{Bayesian Optimisation}
Optimising model hyperparameters is a crucial step in ML models in general. Commonly used methods of searching the hyperparameter space include grid search, random search, and Bayesian optimisation (BO). Grid search samples the hyperparameter space at fixed intervals. Random search tries out a fixed number of random samples of the hyperparameter space. The BO algorithm models the relationship between the hyperparameters and model performance using a probabilistic framework. A Gaussian process surrogate model is fit to the observations of the objective function in each iteration to estimate the optimal point to sample in that iteration. In this method, the knowledge gained by observing previous samples of the hyperparameter space is used to determine the optimal point to sample next, which results in improving the efficiency of searching the hyperparameter space.

To guide the search, BO employs acquisition functions. These functions balance exploration (sampling unexplored regions) and exploitation (focusing on promising areas). In this research, we have utilised expected improvement (EI) as the exploitation function. EI quantifies how much improvement we expect by evaluating a particular hyperparameter configuration. It favours regions where the GP predicts higher function values. BO is an iterative process. The GP is first fit to the observed data. The acquisition function is then optimised to find the next set of hyperparameters. The true objective function (e.g. validation accuracy) is then evaluated with the selected set of hyperparameters. The GP is then updated using the observed results.

\subsection{Computational Complexity}
Denoting the number of input images by $N_I$, the number of pixels in each input image by $N_X$, the number of GQA heads with $H$, the dimensionality of the internal representation of the transformer with $D_A$, and the number of transformer layers with $L$, the computational complexity of the Retformer model is $\mathcal{O}(N_IL(N_X^2D_A+N_XD_A^2+N_X^2H+N_XD_A))$ \cite{vaswani2017attention, keles2023computational, touvron2022three}. The computational complexity of the CNN-based models is $\mathcal{O}(N_ILW^2D_C^2N_X)$ \cite{ke2017relationship}, where $W$ denotes the width of the CNN (the number of feature maps in each layer), $L$ denotes the number of CNN layers, and $D_C$ denotes the convolution dimensionality (kernel size).


\subsection{Explainability}
From the XAI point of view, ML models can be grouped into white-box models, and black-box models \cite{dwivedi2023explainable}. The white-box models are self-explanatory, and the decision-making mechanisms of the model are transparent. The black-box models, on the other hand, are opaque, and their decision-making mechanisms are not clear \cite{dwivedi2023explainable}. The Retformer model and other models used as benchmarks in this research are considered as black-box models \cite{kashefi2023explainability}. While black-box models are not intrinsically explainable like white-box models, there are many different methods that aim to provide explainability for black-box models \cite{kashefi2023explainability, selvaraju2017grad, dwivedi2023explainable}.

A commonly used method of visualising transformer representations is to investigate the attention maps of the transformer. While this approach is a natural choice for transformers and any other neural network that incorporates attention, it also comes with its own limitations. This method can only visualise the attention layers of the network, and it cannot visualise the feature map obtained at other layers, such as the normalisation layer. Furthermore, this method provides a separate visualisation for each attention head in each layer of the network, which does not show the most critical regions of the input image to the overall classification of the transformer.

To alleviate these problems, the Grad-CAM method \cite{selvaraju2017grad} is utilised to visualise the most critical regions of the images for Retformer in AD detection. The Grad-CAM method is based on the observation that in CNNs, spatial feature maps are obtained hierarchically, and the last convolution layer encodes the most important spatial information in the input image \cite{selvaraju2017grad}. This method uses the gradient of each class label with respect to the feature map in the last convolution layer to weight the pixels of the feature map \cite{selvaraju2017grad}. These pixel weights are the output of the Grad-CAM algorithm. 

The key advantage of Gradient-weighted Class Activation Mapping (Grad-CAM) over traditional attention head visualizations lies in its flexibility and depth of interpretability. Grad-CAM allows visualization of activations from any layer or sub-component of the Retformer model, not just the attention heads. This enables inspection of the model’s final encoded representation—just before the classifier head—highlighting the most relevant spatial regions influencing classification decisions. In contrast, attention head visualization is limited to the attention mechanism and typically requires averaging across all attention heads in the final transformer layer. While this provides some insight, it does not capture the effects of subsequent operations such as layer normalization, skip connections, and the SwiGLU activation, which are integral to the model's final decision-making process. Grad-CAM therefore offers a more comprehensive and fine-grained interpretability mechanism for understanding Retformer’s predictions.

\section{Results}
\label{sec.res}
In this research, the following benchmarks are used:
EfficientNet-B3 \cite{corbin2022assessment}, MobileNet \cite{kim2024efficient}, ResNet-50 \cite{wisely2022convolutional}, VGG-16 \cite{yousefzadeh2024neuron}, and the RETFound foundation model for retinopathy detection \cite{zhou2023foundation}. The CNN-based benchmark models (EfficientNet-B3, MobileNet, ResNet-50, and VGG-16) are all initialised from pretrained weights obtained from training on the ImageNet dataset \cite{deng2009imagenet}. This pretraining strategy is also used in previous works \cite{corbin2022assessment, wisely2022convolutional, yousefzadeh2024neuron, kim2024efficient}. The RETFound foundation model is initialised from pretrained weights obtained from training on retinal images \cite{zhou2023foundation}. Early stopping was not used in this study.

\subsection{Hyperparameters}
As explained in Section \ref{ssec.hyperparameters}, nested cross-validation is utilised to train, validate, and test the models, and Bayesian optimisation is employed to search the hyperparameter space. The CNN-based benchmarks (i.e. EfficientNet-B3, MobileNet, ResNet-50, and VGG-16), were all initialised from weights obtained by pre-training on the ImageNet dataset \cite{deng2009imagenet}. The classifier heads are discarded, and dense (MLP) layers are added on top of the bottleneck layer for these benchmarks. The number of dense layers and the number of neurons per layer constitute the hyperparameters of these benchmarks, and are determined during the nested cross-validation procedure. The number of epochs was determined by monitoring the models' convergence over time. We ensured that each model had sufficient time to converge to its final values, which typically occurred within 100 epochs for the Retformer and the VGG-16 models, and within 1000 epochs for the rest of the models. The batch size was set to 250, allowing us to balance computational efficiency with adequate training data exposure.

The RETFound model is simply fine-tuned on our data without any change in the hyperparameters. As a result, it is trained and evaluated on the outer folds of the nested cross-validation. The Retformer model is trained from scratch on the retinal images. The number of transformer layers, number of attention heads and query groups in the GQA, and the number of neurons in the MLP classifier head constitute the hyperparameters of the Retformer model.

The nested cross-validation consists of 3 inner folds and 3 outer folds. As a result, each test subset is comprised of 1/3 of the original data, each validation subset is comprised of 1/9 of the original data, and each train subset is comprised of 4/9 of the original data. 15 iterations of Bayesian optimisation are performed in this research. While we could have manually tuned the hyperparameters based on domain expertise, our goal was to ensure that the selection of hyperparameters was unbiased and data-driven. By leveraging automated tuning, we aimed to identify the most effective combination of hyperparameters for the Retformer model while minimising the risk of overfitting or underfitting due to human bias.

The design of the inner and outer folds in nested cross-validation controls the trade-off between the statistical power of model evaluation and computational complexity. For the sake of simplicity, in this study, we have assumed that the number of inner and outer folds are equal ($K_1 = K_2 = k$). We need to determine minimal $k$ which guarantees desired statistical power (80\%) for the expected effect size (mean/standard deviation of the difference between the accuracy of Retformer and the benchmark models). In this context, the number of degrees of freedom of the hypothesis test involved in model evaluation is given by
\begin{equation}
	DF = k-1,
	\label{eq.df}
\end{equation}
The statistical power of model evaluation is obtained from
\begin{equation}
	P = 1-F_T(t, ES, DF),
	\label{eq.power}
\end{equation}
where $P$ is statistical power, $F_T$ is the cumulative distribution function of student's t-distribution evaluated at $t$, $t$ is the difference between the accuracy of the Retformer model and the best benchmark model, $ES$ is the effect size of the difference between the performance of the Retformer model and the best benchmark model (i.e. mean difference/standard deviation of difference).
Using \eqref{eq.df} and \eqref{eq.power}, it can be seen that using $k=3$ is sufficient to obtain a statistical power of 80\%.

The hyperparameter space for the Retformer model and the CNN-based benchmarks are presented in Table \ref{tab.Retformer_hyperparameter_space} and Table \ref{tab.cnn_hyperparameter_space}, respectively. The prior distributions over the parameters of the Retformer model are all uniform, while the prior distribution over the number of neurons in the MLP classifier head of the CNN-based models is exponential (log-uniform) to prevent overfitting. These prior distributions are sampled in the BO procedure to locate the globally optimal hyperparameter configurations. In other words, the number of neurons in each MLP layer of the CNN-based benchmarks is much more likely to be close to 100 than it is to be close to 4096, and the samples generated in the BO procedure contain more instances close to 100 than instances close to 4096. The hyperparameter ranges are taken from previous studies \cite{corbin2022assessment, wisely2022convolutional, yousefzadeh2024neuron, kim2024efficient}, and extended to ensure the search space is large enough to contain the optimal hyperparameters.

\begin{table}[]
	\centering
	\caption{Retformer Hyperparameter Space}
	\label{tab.Retformer_hyperparameter_space}
	\begin{tabular}{ll}
		\hline
		Hyperparameter                            & Possible Values      \\ \hline
		Projection Dimensionality ($D$)           & \{32, 33, ..., 64\}  \\ 
		Patch Size                                & \{4, 5, 6, 7, 8, 9\} \\ 
		Number of Transformer Layers              & \{1, 2, 3, 4, 5\}    \\ 
		Number of Attention Heads                 & \{1, 2, 3, ..., 25\}    \\ 
		Number of Query Groups   				  & \{1, 2, 3, 4, 5\}    \\ 
		Number of Neurons in the First MLP Layer  & \{32, 33, ..., 64\}  \\ 
		Number of Neurons in the Second MLP Layer & \{16, 17, ..., 32\}  \\ \hline
	\end{tabular}
\end{table}

\begin{table}[]
	\centering
	\caption{Hyperparameter space for CNN-based models}
	\label{tab.cnn_hyperparameter_space}
	\begin{tabular}{lll}
		\hline
		Hyperparameter                  & Possible Values         & Prior Distribution \\ \hline
		MLP Layers            & \{1, 2, 3, 4\}          & Uniform            \\ 
		Neurons per MLP Layer & \{100, 101, ..., 4096\} & Exponential        \\ \hline
	\end{tabular}
\end{table}

The optimal hyperparameters obtained from a simple (non-nested) cross-validation on the entire datasets are represented in Table \ref{tab.cnn_hyperparameters} for the CNN-based benchmarks and Table \ref{tab.Retformer_hyperparameters} for the Retformer model. Note that the number of MLP layers reported for the CNN-based benchmarks includes hidden layers and the output layer. The number of neurons per layer only applies to the hidden layers, as the output layer has a single neuron. It is also noteworthy that the hyperparameter optimisation procedure is done separately for each dataset.

\begin{table}[]
	\centering
	\caption{The hyperparameters of the CNN-based models}
	\label{tab.cnn_hyperparameters}
	\begin{tabular}{lllll}
		\hline
		& \multicolumn{2}{c}{OCT Dataset} & \multicolumn{2}{c}{Fundus Dataset} \\ \hline
		Model           & MLP Layers  & Neurons per Layer & MLP Layers   & Neurons per Layer   \\ \hline
		EfficientNet    & 3           & 829               & 2            & 1335                \\
		MobileNet       & 3           & 870               & 1            & 1                   \\
		ResNet-50       & 3           & 458               & 2            & 351                 \\
		VGG-16          & 4           & 3599              & 4            & 1273                \\ \hline
	\end{tabular}
\end{table}

\begin{table}[]
	\centering
	\caption{Retformer hyperparameters}
	\label{tab.Retformer_hyperparameters}
	\begin{tabular}{lll}
		\hline
		Hyperparameter                            & OCT Dataset & Fundus Dataset \\ \hline
		Projection Dimensionality ($D$)           & 62          & 32 \\ 
		Patch Size                                & 4           & 4  \\ 
		Number of Transformer Layers              & 1           & 4   \\ 
		Number of Attention Heads  				  &	12          & 2   \\ 
		Number of Query Groups  				  &	3           & 1   \\ 
		Number of Neurons in the First MLP Layer  & 47          & 31 \\ 
		Number of Neurons in the Second MLP Layer & 31          & 28 \\ \hline
	\end{tabular}
\end{table}

In order to assess the sensitivity of the Retformer model to hyperparameter configurations, we employed nested cross-validation on 10 different randomly chosen sets of hyperparameters drawn from the prior distribution specified in Table \ref{tab.Retformer_hyperparameter_space}. Table \ref{tab.hyperparameter_statistics_oct} presents the statistics obtained from training Retformer models on the OCT data with these random sets of hyperparameters, and Table \ref{tab.hyperparameter_statistics_fundus} presents the statistics obtained from training Retformer models on the fundus data with these random sets of hyperparameters. It is noteworthy that the random selection has been done separately for the two datasets.

\begin{table}[H]
	\centering
	\caption{Accuracy statistics for Retformer trained on the OCT dataset with 10 different random sets of hyperparameters}
	\label{tab.hyperparameter_statistics_oct}
	\begin{tabular}{ll}
		\hline
		Parameter                   & Value \\ \hline
		Minimum Accuracy            & 85\%  \\
		Maximum Accuracy            & 91\%  \\
		Median Accuracy             & 90\%  \\
		Average Accuracy            & 89\%  \\
		Accuracy Standard Deviation & 2\%  \\ \hline
	\end{tabular}
\end{table}

\begin{table}[H]
	\centering
	\caption{Accuracy statistics for Retformer trained on the fundus dataset with 10 different random sets of hyperparameters}
	\label{tab.hyperparameter_statistics_fundus}
	\begin{tabular}{ll}
		\hline
		Parameter                   & Value \\ \hline
		Minimum Accuracy            & 90\%  \\
		Maximum Accuracy            & 94\%  \\
		Median Accuracy             & 94\%  \\
		Average Accuracy            & 93\%  \\
		Accuracy Standard Deviation & 1\%  \\ \hline
	\end{tabular}
\end{table}

As evident in Table \ref{tab.hyperparameter_statistics_oct} and Table \ref{tab.hyperparameter_statistics_fundus}, the Retformer model is not sensitive to changes in the hyperparameters.

\subsection{Performance Evaluation}
\label{ssec.performance}
The classification quality of the models assessed in terms of accuracy, precision, recall (sensitivity), specificity, F1 score, and AUC on the OCT dataset are tabulated in Table \ref{tab.oct_performance}. The tables present mean $\pm$ standard deviation of the performance metrics. As can be seen, the Retformer model outperforms benchmarks in all classification metrics. 

Table \ref{tab.p_values} presents the p-value of the two sample T-test, testing against the null hypothesis that the difference in the performance of the Retformer model and the second-best performing model (VGG-16) is due to random chance. As evident in Table \ref{tab.p_values}, the differences between the accuracy, precision, sensitivity, specificity, F-1 score, and ROC AUC of the Retformer model and the best benchmark (VGG-16) are all extremely statistically significant, with p<0.001 in all cases.

A number of factors can potentially explain the superior performance of the Retformer model. The Retformer model has fewer parameters than the benchmark models, which avoids overfitting and allows it to generalise better to unseen data.

\begin{table}[h]
	\centering
	\caption{Comparison of classification metrics on the OCT dataset}
	\label{tab.oct_performance}
	\begin{tabular}{lllllll}
		\hline
		Model           & Accuracy                & Precision               & Sensitivity             & Specificity   & F1 Score      & ROC AUC       \\ \hline
		\textbf{Retformer}  & \textbf{92\% $\pm$ 1\%} & \textbf{91\% $\pm$ 2\%} & \textbf{90\% $\pm$ 5\%} & \textbf{91\% $\pm$ 3\%} & \textbf{90\% $\pm$ 4\%} & \textbf{96\% $\pm$ 1\%} \\ 
		RETFound        & 79\% $\pm$ 2\%          & 81\% $\pm$ 2\%          & 73\% $\pm$ 10\%         &  81\% $\pm$ 8\%  & 77\% $\pm$ 6\%  & 79\% $\pm$ 2\% \\ 
		EfficientNet    & 54\% $\pm$ 1\%          & 53\% $\pm$ 2\%          & 62\% $\pm$ 12\%         &  59\% $\pm$ 10\% & 57\% $\pm$ 6\%  & 60\% $\pm$ 2\% \\ 
		MobileNet       & 50\% $\pm$ 2\%          & 26\% $\pm$ 37\%         & 85\% $\pm$ 21\%         &  56\% $\pm$ 17\% & 40\% $\pm$ 53\% & 56\% $\pm$ 6\% \\ 
		ResNet-50       & 73\% $\pm$ 6\%          & 81\% $\pm$ 2\%          & 76\% $\pm$ 10\%         &  80\% $\pm$ 2\% & 78\% $\pm$ 6\%  & 84\% $\pm$ 2\% \\ 
		VGG-16          & 81\% $\pm$ 4\%          & 86\% $\pm$ 1\%          & 80\% $\pm$ 9\%          &  85\% $\pm$ 1\% & 83\% $\pm$ 5\%  & 87\% $\pm$ 2\%         \\ \hline
	\end{tabular}
\end{table}

\begin{table}[h]
	\centering
	\caption{The p-values of the difference between the performance of the Retformer model and the second-best models}
	\label{tab.p_values}
	\begin{tabular}{ll} 
		\hline
		Metric      & p-value \\ \hline
		Accuracy    & 0.0001  \\
		Precision   & 0.0001  \\
		Sensitivity & 0.0001   \\
		Specificity & 0.0001  \\
		F-1 Score   & 0.0002  \\
		ROC AUC     & 0.0001  \\ \hline
	\end{tabular}
\end{table}

Furthermore, the benchmark models have been pre-trained on a large general dataset, and the RETFound model has been pre-trained on a dataset of retinal pathologies. While this can be beneficial, it also means that these models may not have the same level of domain-specific knowledge as a model specifically designed for AD. The Retformer model, on the other hand, being trained from scratch, has learned features that are more relevant to the detection of AD in retinal images. Another reason for the superior performance of the Retformer model in comparison with the CNN-based benchmarks is the ability of the transformer models to learn long-range dependencies in the input image, while the CNN-based models can only learn local dependencies \cite{dosovitskiy2020image}. 

\begin{figure}[H]
	\centering
	\includegraphics[scale=.7]{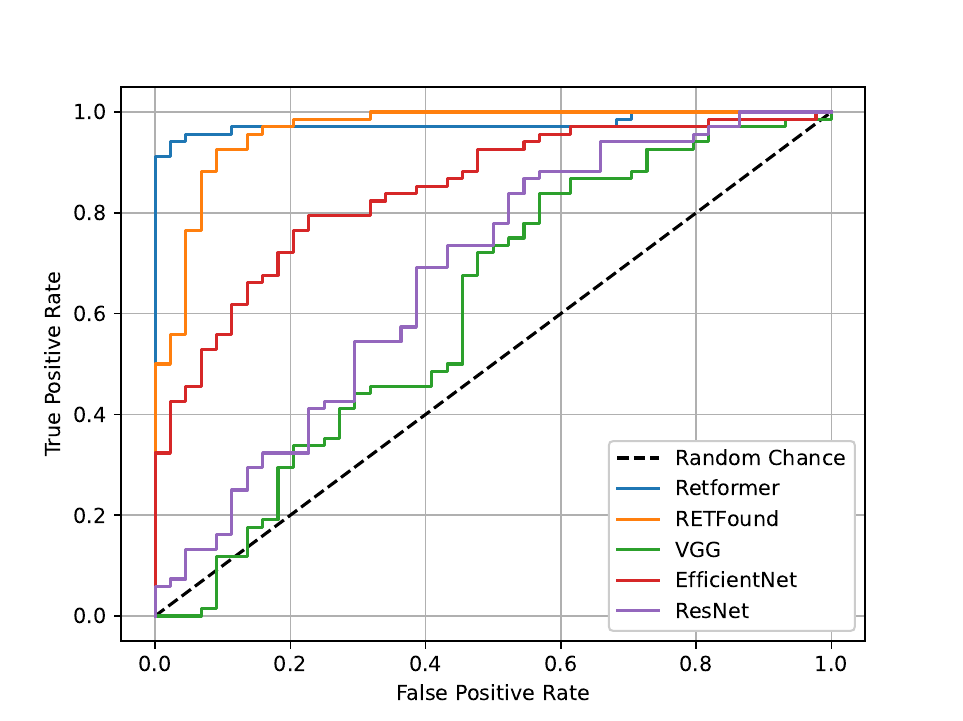}
	\caption{The ROC curve on the OCT data}
	\label{fig.oct_roc}
\end{figure}

The ROC curves obtained by the models are presented in Figure \ref{fig.oct_roc}. The figure shows that the Retformer model matches or outperforms the other models in all thresholds. Also, Table \ref{tab.fundus_performance} presents the classification quality of the models in terms of accuracy and AUC on the fundus dataset. Since this dataset is small, obtaining accurate estimations of the other metrics is difficult. The ROC/AUC curves were generated using standard (non-nested) $k$-fold cross-validation. In this process, each model was initialized with its optimal set of hyperparameters determined during a prior tuning phase. This approach ensures that the reported performance metrics reflect the model’s robustness and generalisability across different data splits.

\begin{table}[b]
	\centering
	\caption{Comparison of classification metrics on the fundus dataset}
	\label{tab.fundus_performance}
	\begin{tabular}{lll}
		\hline
		Model           & Accuracy                & ROC AUC       \\ \hline
		Retformer       & \textbf{94\% $\pm$ 8\%} & \textbf{99\% $\pm$ 3\%} \\ 
		RETFound        & \textbf{94\% $\pm$ 8\%} & 94\% $\pm$ 3\%         \\ 
		EfficientNet    & 89\% $\pm$ 5\%          & 95\% $\pm$ 4\%         \\ 
		MobileNet       & 69\% $\pm$ 3\%          & 66\% $\pm$ 22\%         \\ 
		ResNet-50       & \textbf{94\% $\pm$ 5\%} & 98\% $\pm$ 2\%\\ 
		VGG-16          & 85\% $\pm$ 7\%          & 94\% $\pm$ 7\%   \\ \hline
	\end{tabular}
\end{table}

The VGG-16 outperforms the other benchmarks on the OCT dataset. This could be attributed to the simpler structure of the VGG-16 model in comparison to the ResNet-50, EfficientNet-B3, and MobileNet architectures. While the VGG-16 model has a higher parameter count, it is structurally simpler, and is comprised of 16 convolutional layers. The ResNet-50 model, on the other hand, is comprised of 50 convolutional layers, along with residual connections. The EfficientNet and MobileNet architectures have also been designed to greatly reduce computational complexity. Since this study was not performed under a resource-constrained scenario, these modifications to the generic CNN structure have not introduced any improvement in performance, and have adversely affected performance.
	
The RETFound model, being a vision transformer model, shows robust performance over both datasets and in all metrics. This can be attributed to the capacity of the vision transformer models to learn complex relationships, and to the fact that this model is pre-trained on retinal images, and is better suited to the task of retinal disease diagnosis than the other benchmark models which have been pre-trained on generic image data.
	
Compared to the Retformer model, it can be seen that the benchmark models struggle with capturing the complex patterns and relationships between retinal images and AD  diagnosis. The Retformer model, with its 7.6M parameter count, is less susceptible to overfitting compared to other models like VGG-16 (42M), ResNet-50 (27.5M), and EfficientNet (9.1M), while it is more capable of learning complex relationships between patches of input images and AD diagnosis. We believe that the combination of its smaller parameter count and robustness against overfitting enables the Retformer model to achieve superior performance in classifying retinal images for AD diagnosis.
	
The reduced risk of overfitting, coupled with the Retformer model's ability to learn meaningful representations, allows it to generalise better to unseen data and achieve higher accuracy on the test dataset.

\subsection{Convergence Speed}
\begin{figure*}[h]
	\centering
	\begin{subfigure}{0.48\textwidth}
		\includegraphics[scale=.4]{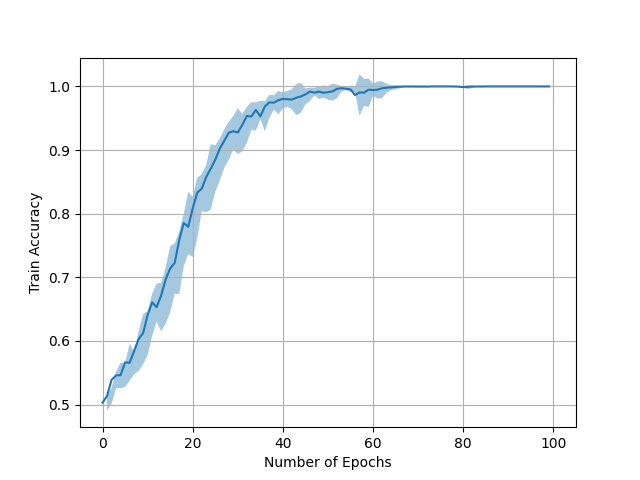}
		\caption{Retformer}
		\label{fig.Retformer_train}
	\end{subfigure}
	\hfil
	\begin{subfigure}{0.48\textwidth}
		\includegraphics[scale=.4]{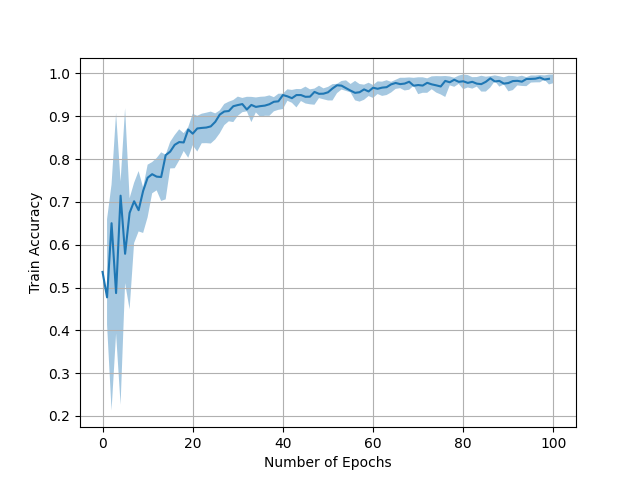}
		\caption{VGG-16}
		\label{fig.vgg_train}
	\end{subfigure}	
	\caption{The training accuracies of the Retformer model and the VGG-16 model on the OCT data}
	\label{fig.train}
\end{figure*}

The training accuracies of the Retformer model and the VGG-16 model (the second-best performing model on the OCT dataset in terms of all metrics other than sensitivity) are compared in this subsection. Figure \ref{fig.train} shows the mean accuracy per epoch, along with one standard deviation highlighted around the mean. To ensure stable convergence, we performed 10 independent runs of the training process, and the average of these runs was reported. We observed no issues with gradients exploding or vanishing during training.

The training accuracy of the Retformer model is presented in Figure \ref{fig.Retformer_train}, showing the speed of convergence of the model. As evident in the figure, the Retformer model converges after 65 epochs. Figure \ref{fig.vgg_train} shows the training accuracy of the VGG-16 model. The figure shows that the VGG-16 model converges after around 100 epochs. Remember that the VGG-16 model is being fine-tuned while the Retformer model is being trained from scratch. This shows the superior convergence speed of the Retformer model, which is also observed by \cite{dosovitskiy2020image}. It is also noteworthy that the VGG-16 model, along with the optimal classifier head obtained in hyperparameter tuning, has 42 million parameters, while the Retformer model, with its optimal set of hyperparameters, has 4 million parameters, showcasing how the Retformer model can achieve comparable performance with fewer parameters.

Furthermore, the convergence plots in figure \ref{fig.train} demonstrate that the Retformer model not only converges faster but also exhibits improved training stability, as evident from the narrower confidence intervals of the train accuracy curves. This suggests that the Retformer model is less prone to overfitting or underfitting during training, which can lead to better generalization performance on unseen data.

\subsection{Downsampling}
\label{ssec.down_sampling}
\begin{figure}[H]
	\centering
	\includegraphics[scale=0.65]{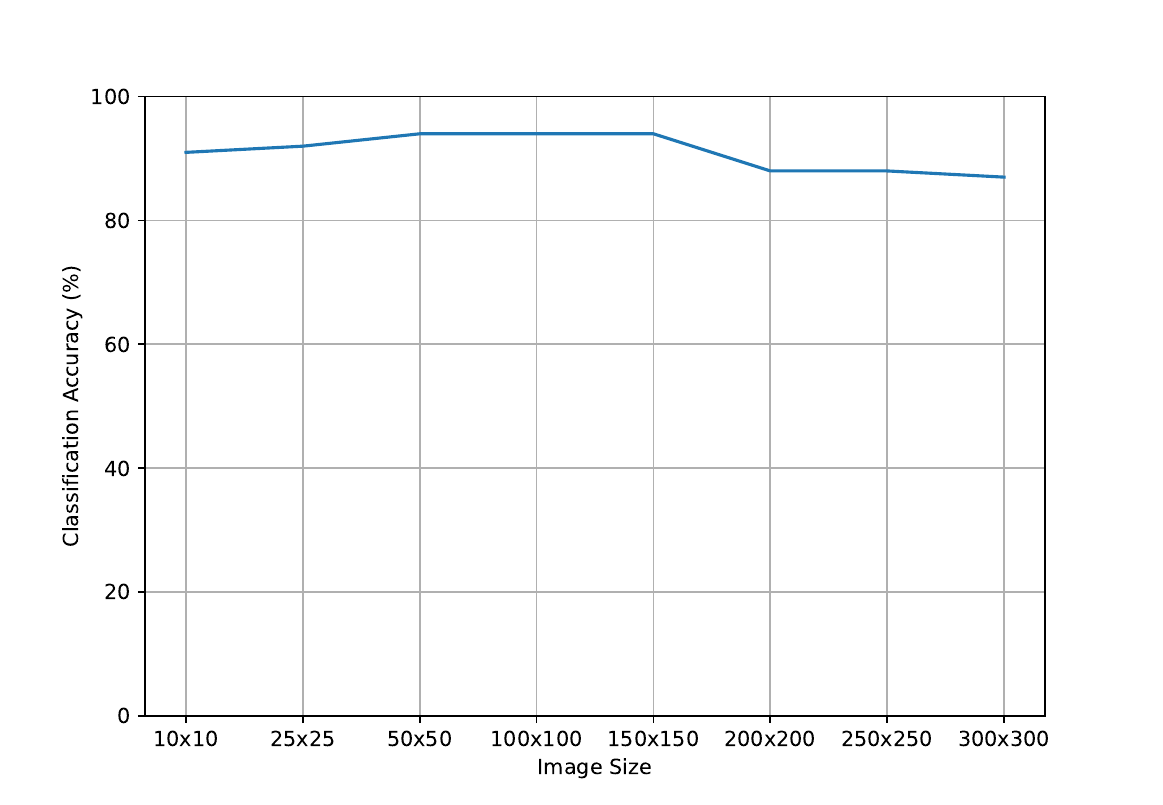}
	\caption{Accuracy vs. Image Size}
	\label{fig.downsampling}
\end{figure}

To address the potential impact of downsampling on our results, we conducted experiments to evaluate the effect of varying image sizes on model performance. We trained the Retformer model on downsampled images of sizes ranging from 10x10 to 300x300 pixels. The results can be seen in figure \ref{fig.downsampling}. These results are obtained on the fundus dataset.

As evident in the figure, by increasing input image size, model accuracy initially improves due to the increase in the information contained in each image. Afterwards, accuracy plateaus for a while, and afterwards increasing image size leads to overfitting and decreased performance, as the number of images stays constant while the number of features (pixels) in each image increases, leading to an increase in the probability of overfitting.

\subsection{Batch Size Sensitivity}
\label{ssec.batch_size}
\begin{figure}[H]
	\centering
	\includegraphics[scale=0.5]{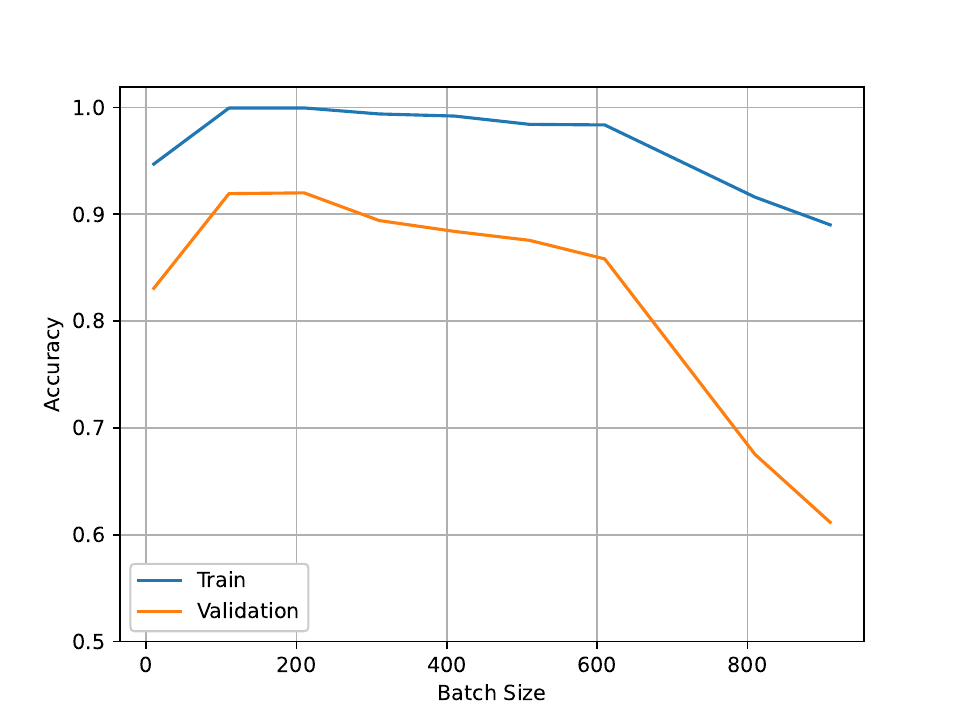}
	\caption{Training and Validation Accuracy vs. Batch Size}
	\label{fig.batch_size}
\end{figure}

The train and validation performances of the model with different batch sizes are presented in Fig. \ref{fig.batch_size}. As evident in the figure, the validation performance of the model peaks around batch sizes of 100 to 300, and drops afterwards.

\subsection{Ablation Study}
\label{ssec.ablation}
An ablation study is undertaken in this research to determine the contribution of each building block of the Retformer model to the overall classification, and to compare the impact of the blocks used in the Retformer with the blocks of the original vision transformer architecture \cite{dosovitskiy2020image}.
This ablation study is performed on the OCT dataset, with the optimal hyperparameters presented in Table \ref{tab.Retformer_hyperparameters}. Here, we have utilised a simple 3-fold cross-validation instead of nested cross-validation to determine the accuracy of the model resulting from removing each building block of the transformer, since the hyperparameters are known beforehand. The results are presented in Table \ref{tab.ablation}.

The results show the immense importance of the first residual connection that connects the transformer layer's input with the GQA output. Furthermore, it can be seen that RoPE significantly outperforms learned position embeddings in this study. It is also evident that using MLP heads instead of convolutional layers to embed the input patches, or instead of SwiGLUs to obtain the transformer representations adversely affects performance. It can be seen that using multihead attention instead of GQA is even worse than removing the attention block altogether. While this might seem counterintuitive, it has been observed that vision transformers can learn reasonably well without attentions \cite{wang2022shift}.
On the other hand, removing the first layer normalisation block does not affect the performance of the model greatly.

\begin{table}[]
	\centering
	\caption{Ablation Results}
	\label{tab.ablation}
	\begin{tabular}{ll}
		\hline
		Change                                & Model Accuracy \\ \hline
		None (Baseline)                          & 94\%       \\ 
		Using learned position embeddings        & 80\%       \\ 
		Using MLP for patch embedding            & 89\%       \\ 
		First layer normalisation removed        & 92\%       \\ 
		Using multihead attention instead of GQA & 85\%       \\ 
		Attention head removed                   & 89\%       \\ 
		First residual connection removed        & 57\%       \\ 
		Second layer normalisation removed       & 90\%       \\ 
		Using MLP with GELU instead of SwiGLU    & 86\%       \\ 
		SwiGLU removed                           & 89\%       \\ 
		Second residual connection removed       & 90\%       \\ \hline
	\end{tabular}
\end{table}

\subsection{XAI}
\label{ssec.xai}
\begin{figure*}[h]
	\centering
	\begin{subfigure}{0.22\textwidth}
		\includegraphics[scale=.7]{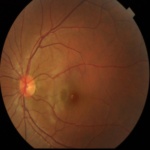}
		\caption{A visible light fundus image}
		\label{fig.fundus}
	\end{subfigure}
	\hfil
	\begin{subfigure}{0.22\textwidth}
		\includegraphics[scale=.7]{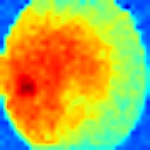}
		\caption{Feature map visualised using Grad-CAM}
		\label{fig.map}
	\end{subfigure}
	\hfil
	\begin{subfigure}{0.22\textwidth}
		\includegraphics[scale=.7]{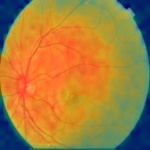}
		\caption{The output of the Grad-CAM algorithm}
		\label{fig.cam}
	\end{subfigure}
	\hfil
	\begin{subfigure}{0.22\textwidth}
		\includegraphics[scale=.7]{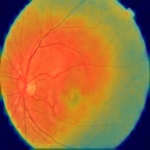}
		\caption{The average of all attention heads}
		\label{fig.mean_attention}
	\end{subfigure}		
	\caption{Visual explanation for Retformer classification of visible light fundus images.}
	\label{fig.fundus_cam}
\end{figure*}

Figure \ref{fig.fundus_cam} displays the Grad-CAM output for the Retformer classifier and visualisation of its attention heads, explaining the input image's classification. Visualising the feature map obtained by the last layer normalisation for the fundus image in Figure \ref{fig.fundus} using Grad-CAM results in the colour map presented in Figure \ref{fig.map}. In this image, red represents the largest weights and blue represents the smallest weights. Combining Figure \ref{fig.fundus} with Figure \ref{fig.map} results in Figure \ref{fig.cam}, which offers a visual explanation of how the Reformer model classifies the visible light fundus images.

As evident in Figure \ref{fig.cam}, the retinal vasculature on the left side of the fundus image and the optic nerve head (optic disc), which is also towards the left of the image, have the highest weights in distinguishing AD patients from HCs, and the right side of the fundus image has a lower weight. This is in line with clinical studies, which connect AD with changes in the retinal vasculature \cite{cheung2021retinal, cheung2017imaging, lee2020retinal}. Clinical studies have also demonstrated correlations between the size \cite{budu2020relationship} and colour \cite{BAMBO201568} of the optic disc and AD.

\begin{figure*}[h]
	\centering
	\begin{subfigure}{0.24\textwidth}
		\scalebox{2.5}[0.75]{
			\includegraphics[angle=90, scale=.32]{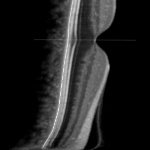}
		}
		\caption{OCT image slice}
	\end{subfigure}
	\hfil
	\begin{subfigure}{0.24\textwidth}
		\scalebox{2.5}[0.75]{
			\includegraphics[angle=90, scale=.32]{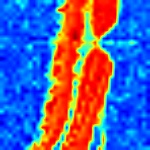}
		}
		\caption{Feature map}
	\end{subfigure}
	\hfil
	\begin{subfigure}{0.24\textwidth}
		\scalebox{2.5}[0.75]{
			\includegraphics[angle=90, scale=.32]{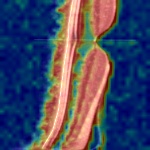}
		}
		\caption{Grad-CAM output}
	\end{subfigure}
	\hfil
	\begin{subfigure}{0.24\textwidth}
		\scalebox{2.5}[0.75]{
			\includegraphics[angle=90, scale=.32]{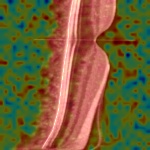}
		}
		\caption{Average attention}
	\end{subfigure}
	\caption{Visual explanation for Retformer classification of OCT images.}
	\label{fig.oct_cam}
\end{figure*}

Figure \ref{fig.mean_attention} presents a visualisation of the attention heads of the Retformer model trained to classify fundus images. While this visualisation shows the importance of the retinal vasculature and the optic disc, the Grad-CAM output offers a better visual explanation and visualises a layer closer to the output of the Retformer model than the GQA. Figure \ref{fig.oct_cam} shows the Grad-CAM output and the average of the attention heads for a slice of an OCT image. As evident in Figure \ref{fig.oct_cam}, the thickness of the retina, including RNFL, the ganglion layer, and nuclear layers, along with the choroid layer, have the greatest weights in distinguishing AD patients from HCs. The choroid layer is responsible for the retina's blood supply and has one of the highest blood flows per volume in the entire human body \cite{cheung2021retinal}. Clinical studies suggest a relationship between the thickness of the retina and AD \cite{cheung2021retinal, cheung2017imaging, den2017retinal, kirbas2013retinal} as well. They also suggest a relationship between changes in the thickness of the choroid layer and AD \cite{cheung2021retinal, cheung2017imaging, lee2020retinal}.
Since both visible light fundus images and OCT slices show the importance of the retinal vasculature and blood flow in AD detection, future work may focus on OCTA images specifically targeting the vasculature.

Ophthalmologist-labelled data was not available to facilitate a comparison with the Grad-CAM algorithm. However, in order to quantitatively assess the explainability of the Retformer model, we designed the following experiment. We employed nested cross-validation to ensure the reliability of our results, as it allows us to assess the generalisability of our findings while controlling for overfitting.
	
	In this setup, we trained a Retformer model on the training data of the outer folds of the nested cross-validation, and used it to generate regions of interest in the test images of the outer folds. We then created two types of masked images: one where the regions of interest are selected using Grad-CAM and the rest of the image is masked, and another with randomly generated masks. These images, along with the original images, allow us to compare the performance of the model when relying solely on the regions of interest detected by Grad-CAM with the performance of the model trained on masked data using random masks, and with the performance of the model trained using the original data.
	
	We split the test data of the outer folds data into a train and a test subset in the inner folds, and trained three different randomly initiated Retformer models with the optimal hyperparameters: one on the original data, one on the data masked by Grad-CAM, and one on the randomly masked data. The nested cross-validation setup involved 2 inner folds, 2 outer folds, and 25 different randomly initiated repetitions of the whole setup.
	
	The accuracies of the models were then compared using two-sample t-tests with a significance threshold of 0.025, corresponding to a significance threshold of 0.05 corrected for repeated measurements using the Bonferroni correction.
	
	Our results show that the accuracy of the model trained on the original data and the accuracy of the model trained on the data masked by Grad-CAM did not differ significantly (p>0.05), indicating that the regions of interest detected by Grad-CAM are indeed relevant for AD diagnosis. In contrast, the model trained on the data masked by Grad-CAM and the model trained on the randomly masked data differed significantly (p<0.001). This suggests that the regions of interest detected by Grad-CAM contribute meaningfully to the model's performance.
	
	By demonstrating the effectiveness of Grad-CAM in explaining model predictions, we provide evidence supporting the use of this technique in interpreting the results of our Retformer model. This enhances the transparency and reliability of our findings, allowing for more informed decision-making in the context of AD screening and diagnosis using retinal imaging modalities.

\section{Discussion}
\label{sec.disc}
In this research, we propose Retformer, a transformer-based neural network architecture for AD detection using retinal images. The transformer-based architecture of Retformer is capable of learning long-range dependencies in the images, as opposed to the local dependencies learned by CNN-based models. Furthermore, Every module in the Retformer model is designed to reduce overfit and improve performance. The Retformer model is trained on domain-specific data, i.e. retinal images from AD patients and HC individuals. As a result of these design choices, the Retformer model outperforms state-of-the-art benchmarks \cite{corbin2022assessment, wisely2022convolutional, yousefzadeh2024neuron, kim2024efficient, zhou2023foundation} in terms of accuracy, precision, specificity, sensitivity, and AUC, with an accuracy of 92\%, outperforming the best benchmark by a margin of 11\%, as shown in Subsection \ref{ssec.performance}. The impressive performance of the Retformer model over the smaller fundus datasets highlights the capability of this model to be trained on a relatively small, task-specific dataset and provide robust classification with lower computational cost, making the Retformer an attractive choice for AI-aided clinical diagnosis.

We designed the Retformer architecture with several motivations in mind. Firstly, our primary goal was to leverage the strengths of transformers in modelling long-range dependencies within the retinal image data. Given the relatively small size of our training dataset, we also sought to minimise the risk of overfitting while maximising the model's ability to generalise to unseen samples. The Retformer model can be considered a hybrid CNN-transformer model, as it leverages convolutional filters in its patch encoder to encode the visual data in each patch of the input image, resulting in better performance in comparison to traditional transformer models, as shown in the ablation study conducted in Subsection \ref{ssec.ablation}.

While the optimal set of the hyperparameters of the Retformer model was obtained using Bayesian optimisation, it was shown in \ref{ssec.hyperparameters} that the Retformer model is not sensitive to hyperparameter configurations, and the standard deviation of the accuracy obtained using 10 different random sets of hyperparameters was much smaller than the margin between the performance of the Retformer model and the best benchmark model (2\% vs. 11\%).

As shown in \ref{ssec.batch_size}, the impact of batch sizes on the performance of the Retformer model is dictated by the bias-variance tradeoff. Smaller batch sizes result in higher variance, while larger batch sizes result in higher bias, both degrading the performance of the model. Batch sizes of 100-300 samples result in striking a balance between bias and variance errors, and optimal performance. The same principle is true for down sampling, where small images sizes result in increased bias while larger image sizes result in increased variance, with a sweet spot between image sizes of 50x50 and 150x150 pixels.

The ablation study presented in Subsection \ref{ssec.ablation} validates the design choices in the Retformer model. The RoPE encoder is able to provide informative positional encoding without adding more trainable parameters to the model and outperforms the trainable position encoding scheme. The CNN-based patch encoding mechanism utilised in the Retformer model is better suited to the task than the standard MLP patch encoding used in the original vision transformer model. The GQA attention head of the Retformer model is much more efficient than the multi-head attention mechanism used in the original vision transformer. The SwiGLU activation function produces more informative features than the GELU activation function, and noticeable improves the performance of the model.

Visualising the decision-making process of the Retformer model using the Grad-CAM algorithm suggest the importance of the retinal vasculature and blood flow in distinguishing AD patients from HC individuals. This finding, presented in Subsection \ref{ssec.xai}, aligns well with clinical research, and validates the meaningfulness of the representations learned by the Retformer model. It also shows that XAI in general, and the Retformer and Grad-CAM pipeline in particular, offer the potential of providing valuable insights into the underlying mechanisms of diseases such as AD and the changes they cause in human physiology and anatomy.

The proposed Retformer architecture has significant implications for the early detection and diagnosis of Alzheimer's disease by providing a non-invasive, cost-effective, and efficient means of identifying individuals at risk. The clinical implications of our findings are multifaceted:

\begin{itemize}
	\item \textbf{Early intervention}: By detecting AD-related changes in retinal morphology, clinicians can initiate early interventions, such as lifestyle modifications, cognitive training programs, or pharmacological treatments, which may slow disease progression.
	\item \textbf{Personalized medicine}: The Retformer's ability to analyse individual retinal images enables personalised plans tailored to each patient's specific needs and risk factors.
	\item \textbf{Improved diagnostic accuracy}: The Retformer model's high sensitivity and specificity can reduce false positives and negatives, leading to more accurate diagnoses and better patient outcomes.
\end{itemize}
However, several limitations must be considered when applying the Retformer to real-world screening scenarios:
\begin{itemize}
	\item \textbf{Data quality and availability}: The performance of the Retformer relies heavily on high-quality labelled retinal image datasets. Ensuring access to diverse, representative datasets will be crucial for widespread adoption.
	\item \textbf{Clinical validation}: Further studies are needed to validate the Retformer's performance in clinical settings, including assessing its effectiveness in detecting AD in various populations and evaluating its integration with existing diagnostic workflows.
	\item \textbf{Regulatory frameworks}: As XAI-driven solutions like the Retformer are developed for use in medical settings, regulatory bodies will need to establish guidelines and standards for their deployment, ensuring compliance with data protection, patient confidentiality, and healthcare regulations.
\end{itemize}

Our study may be limited by the relatively small size of the datasets used. However, since we did not have access to any other openly available dataset of retinal images from AD patients, we used the only openly available datasets. While cross-validation helps mitigate this issue, we recognise that external validation on independent datasets would provide stronger evidence for our findings. We were unable to access additional datasets for further validation purposes.

The datasets we have used in this research may contain potential biases that could affect the model's generalisability.
Firstly, demographic and geographical biases are possible, as each of the two datasets contains data from a limited number of participants.
Additionally, device-related biases may exist since the OCT dataset is collected using a specific OCT scanner, and the fundus dataset is collected using a fixed fundus camera.
Selection bias is also possible, as published datasets only include patients who consented to participate. We aim to address this by recruiting a more diverse population in future datasets.
However, we have made efforts to balance these aspects by ensuring that HCs and AD patients are of similar ages, which is why we have discarded the data from the young adult controls in our study. The published datasets also have a similar gender distribution in both AD and HC groups, ensuring that the results are not gender biased.

Overall, this research has significant implications for the fields of ophthalmology and clinical neuroscience, and provides XAI-powered diagnosis aid and valuable insight into the retinal biomarkers of AD, enabling screening, early detection, and better management options for AD patients. Based on our findings, future research may focus on leveraging OCTA to further explore the link between the degenration of retinal vasculature and AD. The Retformer model could also be utilised in other medical image analysis tasks, including brain MRI, CT and PET scans, and XAI options could be explored in medical settings, in general, to aid clinicians in providing a better understanding of the impact of various diseases on human anatomy and physiology.
\section{Conclusion}
\label{sec.con}
We proposed Retformer, a novel transformer-based model for retinal image classification. We compared multiple classification algorithms in detecting AD using retinal visible light fundus images and OCT image slices using a number of classification metrics including accuracy and AUC. The Retformer classifier outperformed the other models in all classification performance metrics, offering an improvement of over 11\% against the second-best model in terms of accuracy, and over 5\% improvement in preceision, sensitivity, and specificity. The Grad-CAM algorithm was utilised to visualise the most critical regions in retinal images for AD detection, highlighting the importance of retinal vasculature, the choroid layer, and also the macula and the overall thickness of the retina. Finally, the alignment of these findings with clinical studies was assessed. 

\section*{Data Availability}
The datasets used in this study are available from the Zenodo and Dryad repositories, https://zenodo.org/records/7678656, https://datadryad.org/stash/dataset/doi:10.5061/dryad.msbcc2ftc.

\section*{Competing Interests}
The authors declare no competing interests.

\section*{Acknowledgment}
Farshid Hajati and Raymond Chiong would like to acknowledge funding support from the University of New England, Australia, under Grant No. A24/2676.

\bibliography{elsarticle-template-num}


\end{document}